\documentclass[10pt,twocolumn,letterpaper]{article}

\usepackage{cvpr}              

\definecolor{cvprblue}{rgb}{0.21,0.49,0.74}
\usepackage[pagebackref,breaklinks,colorlinks,allcolors=cvprblue]{hyperref}
\usepackage{hyperref}       
\usepackage{url}            
\usepackage{booktabs}       
\usepackage{amsfonts}       
\usepackage{nicefrac}       
\usepackage{microtype}      
\usepackage{xcolor}         
\usepackage{hyperref}       
\usepackage{url}            
\usepackage{booktabs}       
\usepackage{amsfonts}       
\usepackage{nicefrac}       
\usepackage{microtype}      
\usepackage{xcolor}         
\usepackage{amsmath}
\usepackage{graphicx}
\usepackage{amsmath}
\usepackage{algorithm}
\usepackage{algorithmic}
\usepackage{amsfonts}
\usepackage{multirow}

\def\paperID{14} 
\def\confName{CVPR}
\def\confYear{2025}

\title{Coordinated Robustness Evaluation Framework for Vision-Language Models}

\author{
  Ashwin Ramesh Babu\thanks{Equal Contribution}, Sajad Mousavi$^*$, Vineet Gundecha, Sahand Ghorbanpour, \\
  Avisek Naug, Antonio Guillen, Ricardo Luna Gutierrez, Soumyendu Sarkar\thanks{Corresponding Author}  \\ \\
  Hewlett Packard Enterprise (Hewlett Packard Labs)\\
  \texttt{\{ashwin.ramesh-babu, sajad.mousavi, vineet.gundecha} \\ 
  \texttt{sahand.ghorbanpour, avisek.naug, antonio.guillen, rluna} \\
  \texttt{soumyendu.sarkar\}@hpe.com} \\
}
\begin{document}
\maketitle

\begin{abstract}
Vision-language models, which integrate computer vision and natural language processing capabilities, have demonstrated significant advancements in tasks such as image captioning and visual question and answering. However, similar to traditional models, they are susceptible to small perturbations, posing a challenge to their robustness, particularly in deployment scenarios. Evaluating the robustness of these models requires perturbations in both the vision and language modalities to learn their inter-modal dependencies.  In this work, we train a generic surrogate model that can take both image and text as input and generate joint representation which is further used to generate adversarial perturbations for both the text and image modalities. This coordinated attack strategy is evaluated on the visual question and answering and visual reasoning datasets using various state-of-the-art vision-language models. Our results indicate that the proposed strategy outperforms other multi-modal attacks and single-modality attacks from the recent literature.  Our results demonstrate their effectiveness in compromising the robustness of several state-of-the-art pre-trained multi-modal models such as instruct-BLIP, ViLT and others.

\end{abstract}

\section{Introduction}
Evaluating the robustness of computer vision models and architectures has existed for a long time now.  The success of vision-language models in bridging the gap between visual and textual representations has enabled a wide range of applications, from image captioning, visual question and answering, multi-modal information retrieval and generation tasks \cite{dai2024instructblip, liu2024llava, kim2021vilt}.  As these models are becoming more ubiquitous in real-world deployments, their vulnerability to adversarial attacks poses a significant liability concern. These attacks can deceive models into misinterpreting or misclassifying visual and textual inputs, leading to erroneous outputs and potentially harmful consequences. Understanding and mitigating these threats is crucial to ensuring the reliability and security of vision-language models as they become more integrated into critical applications.  \\
Adversarial examples, carefully crafted by adding imperceptible perturbations to input data, can cause these models to make incorrect and potentially dangerous predictions.  While adversarial attacks have been extensively studied in computer vision and natural language processing domains, the unique challenges posed by the multimodal nature of the vision-language model necessitate more detailed investigation.  These models must contend with adversarial perturbations that can manifest in both the visual and textual components simultaneously, potentially exploiting intricate cross-modal interactions and challenging the model's ability to reason coherently across modalities.  Furthermore, most of the studies in the computer vision domain and natural language processing (NLP) are designed only for classification tasks, Vision and Language models mainly involve different types of downstream tasks such as visual question and answering, and cross-modality retrieval.

\begin{figure*}[h]
    \centering
    \includegraphics[scale=0.70]{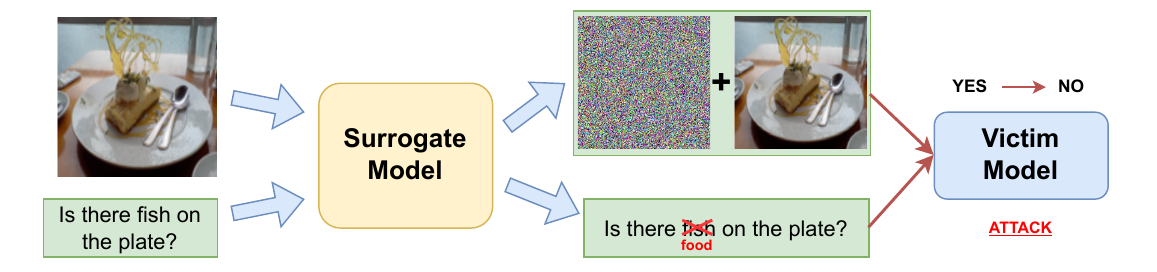}
    \caption{Sample outcome with the proposed method on VQA dataset model with ViLT as victim model}
    \label{fig:introduced}
\end{figure*}

\begin{figure*}[h]
    \centering
    \includegraphics[scale=0.70]{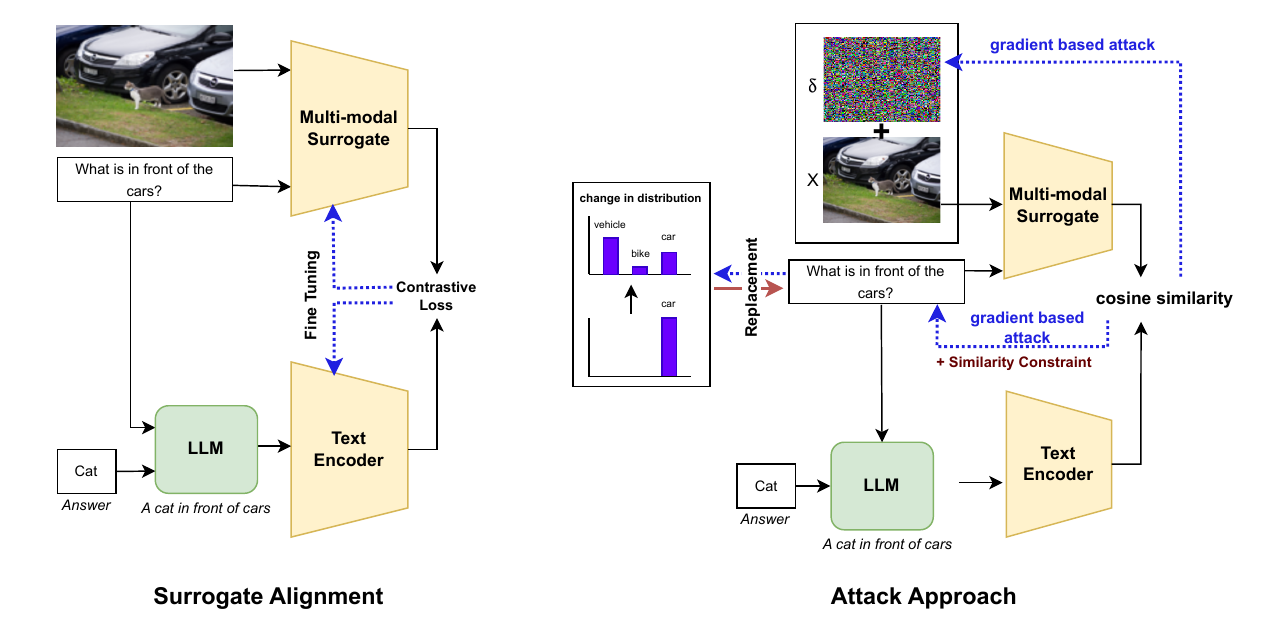}
    \caption{Overview of the workflow.  The left figure represents the alignment step to align a custom architecture comprising of an image and question module with the answer module.  An additional LLM component is used in the alignment step to convert a single-worded ground-truth to a sentence based on the question.  The dotted line represents the flow of gradients. The figure in the right represents the attack approach.  The embeddings generated by the multi-modal surrogate and the text encoder are compared using cosine similarity and are used to generate perturbations for both the text and the image modality. The figure represents one iteration of the attack and the training process.  The surrogate once aligned is used across all victim models which proves the generalizability of the proposed method. }
    \label{fig:main}
\end{figure*}

In this work, we propose a novel coordinated attack strategy that introduces perturbations to both the vision and language modules, aiming to cause the model to alter its outcomes. Our approach involves several key steps: \\
First, we train a multi-modal encoder (Multi-modal surrogate in figure \ref{fig:main} left) to align the embeddings generated from a combination of an image and the corresponding question with those generated by an answer encoder (text encoder in figure \ref{fig:main} left). Specifically, the multi-modal encoder processes an image and a corresponding question, generating a joint representation. This joint representation is encouraged to closely match an answer representation produced by a transformer-based text encoder. The trained multi-modal encoder then serves as a surrogate for our attack method.  During the attack phase, both the surrogate and the text encoder generate feature representations. The goal is to craft adversarial perturbations that push these representations apart in the embedding space, achieved through a gradient-based update for both image and text modalities. This process is formulated as an optimization problem.  Figure \ref{fig:introduced} shows a sample output from our proposed method. Our approach differs from existing methods in several significant ways:

While most literature on multi-modal attacks deals with classification problems where output logits are readily accessible, our method targets complex downstream tasks such as visual question and answering (VQA) rather than simple classifications.  Additionally, majority of surrogate-based approaches use feedback from victim models to generate adversarial perturbations. our method employs a generic surrogate model that generates adversarial samples that are not victim model dependent and can effectively mislead several state-of-the-art victim models.  Finally, most multi-modal adversarial attack approaches in the literature use word-swapping techniques \cite{li2020bert} for text attacks, which result in uncoordinated changes between text and image modalities. In contrast, our method ensures that perturbations in both modalities are coordinated, enhancing the effectiveness of the attack.  The contribution of our approach can be summarized as;

\begin{enumerate}
    \item We propose a coordinated attack strategy that has been designed for both vision and language components to highlight the unique vulnerabilities in the multi-modal context. 

    \item The proposed method acts as a surrogate model that can craft generic adversarial samples that can be used against several victim models without model specific feedback. 
    
    \item Results on visual question and answering task and visual reasoning task demonstrate the superiority of the proposed method when compared to our competitors. 

\end{enumerate}

\section{Related Works}

Adversarial attacks were first introduced in computer vision, which demonstrates the vulnerability of neural networks.  Unlike black-box attacks, where the attacker has no access to the model's internal parameters, white-box attacks assume full knowledge of the model, allowing for more precise and effective perturbations. One of the pioneering works in this area is by \citet{szegedy2013intriguing}, who demonstrated the vulnerability of neural networks to adversarial examples in the context of image classification. This concept was later extended to text models by \citet{papernot2016crafting}, who introduced a technique for generating adversarial sequences by leveraging the gradients of the model.

\subsection{Attacks for text models}
Subsequent works have built upon these foundations, exploring various methods for crafting adversarial text to mislead text based models \cite{xu2020adversarial, wang2021measure, wang2019towards, zhang2020adversarial, zhang2020adversarial}.  Adversarial attack on text are broadly divided into two approaches, where one deals with sentence level and the other at word level. The primary difference between the sentence-level attacks and the word-level attacks is in the granularity and the nature of the modification made to the input text.  The sentence level attack involve modifying entire sentences or adding new sentences to the text \cite{wang2020cat, lin2021using, huang2021generating, han2020adversarial}

\subsubsection{Word-level attacks}
Word level adversarial attack focus on modifying individual word within a sentence.  This attack deals with smaller and granular changes. Some of the most impactful works in this area include Bert-attack \cite{li2020bert}.  Jin et al. in the work \cite{jin2020bert} propose a textfooler which uses a powerful pretrained BERT to generate adversarial adversarial attack. Some of the popular attacks in the word level \cite{he2021model}
Yang et al. in their work \cite{yang2022natural} proposes adversarial attack that is capable of adversarially transforming inputs to make victim models produce wrong outputs. Similarly, there have been deveral other derivatives of the listed attacks such as \cite{garg2020bae, sun2020adv, xu2021r,li2020contextualized, ye2022texthoaxer, chang2023textguise}.  

\subsection{Attacks for vision models}
Adversarial attacks on vision models have been studied for over a decade.  Some of the significant reviews on adversarial attack for vision models are \cite{akhtar2021advances, xu2020adversarial, khamaiseh2022adversarial, chakraborty2021survey}.  Among adversarial attacks, square attack has created significant impact in the adversarial attack literature \cite{andriushchenko2020square}.  Similarly, \cite{moosavi2016deepfool} proposes an efficient approach to fool deep networks using gradient information. The authors use a natural saddle point formulation to capture the notion of security against adversarial attacks in a principled manner.  In the past couple of years, the approach of using reinforcement learning for adversarial attack on image models and explainability has increased \cite{sarkar2024benchmark, sarkar2025RLAB, tsingenopoulos2019autoattacker, sarkar2024robustness, 10260607, sarkar2023robustness, sarkar2023robustnessa, sarkar2022measuring, sarkar2023rl}.

\subsection{Multi-modal Adversarial attacks}
With recent developments in the multi-modal foundational models, they have similar vulnerabilities as the text models or computer vision models.  Zhou et al. \cite{zhou2024revisiting} in their work introduce a multimodal attack to align clean and adversarial text embeddings with clean and adversarial visual features.  They evaluate their method on image classification tasks to prove their superiority.  Zhang et al. \cite{zhang2022towards} in their work co-attack add perturbations on multi-modality settings.  The work uses CLIP-based architecture which are aligned to generate adversarial perturbations for the image and text modality. The authors evaluate their work on image text retrieval and visual entailment tasks to demonstrate the effectiveness of their work. Zhao et al. \cite{zhao2024evaluating} in their work craft adversarial examples to fool VLMs for image captioning tasks by alternating between text-to-image and image-to-text models with having a specific target text to drive the perturbation towards a specific direction.  VLAttack proposes an attack strategy which involves querying a black-box model exhaustively to learn the mutual connections between the perturbed image and text to cause misclassification in their work \cite{yin2023vlattack}.

\section{Problem Formulation}

In this work, we aim to generate adversarial perturbations for an image and a text question to make their combined representation and the corresponding answer representation move apart in the embedding space. These adversarial samples to cause different responses from a variety of victim models compared to the original inputs.  Let $x_q, x_i, x_a$ be a sample from dataset $D$. 
\begin{align*}
    (x_q, x_i, x_a) \sim \mathcal{D}
\end{align*}
$E_{iq}(x_i, x_q) = \mathbf{r}_{iq}$ Encodes an image $i_m$ and a question $i_q$ into a joint representation $r_{iq}$.  $E_a(x_a) = \mathbf{r}_a $ encodes an answer $x_a$ into a representation $r_a$.
Generate perturbations $\delta x_i$ for the image and $\delta x_q$ for the question such that the similarity between $\mathbf{r}_{iq}$ and $\mathbf{r}_{a}$ decreases, pushing $ \mathbf{r}_{iq} $ and $\mathbf{r}_a $ farther apart in the embedding space.  The perturbed image and question should have different responses from a victim model compared to the original inputs. We formulate the problem as an optimization task where we minimize the similarity between the perturbed joint representation and the answer representation while also causing different outputs from a victim model $V$.

Let \( V(x_i, x_q) \) be the response of the victim model to the original image and question and $ V(x_i + \Delta x_i, x_q + \Delta x_q) $ be the response of the victim model to the perturbed image and question.We want $ V(x_i + \Delta x_i, x_q + \Delta x_q) \neq V(x_i, x_q) $.

So, the overall problem can be formulated as, 
\[ \min_{\Delta x_i, \Delta x_q} \text{similarity}(E_{iq}(x_i + \Delta x_i, x_q + \Delta x_q), E_a(x_a)) \]
\[ \text{subject to} \quad V(x_i + \Delta x_i, x_q + \Delta x_q) \neq V(x_i, x_q) \]

\section{Proposed Method}
In the proposed method, we first explain the architecture of the surrogate model used and how the surrogate model was aligned for the purpose of the adversarial example generation.  Next, we explain the attack strategies for the individual modalities, and how they are combined to generate adversarial perturbation.  The overall flow of the proposed method is represented in the figure \ref{fig:main}.

\subsection{Surrogate Model to generate perturbations}

\subsubsection{Surrogate architecture}

Current approaches in vision-language models heavily rely on image feature extraction processes that involve regional supervision such as object detection which is computationally more expensive and has limitations in applications. Hence at the core of the surrogate model lies a transformer-based architecture to capture the interaction between the text and the image to generate a joint representation. The surrogate architecture handles two modalities in a single unified manner consisting of stacked blocks that include a multithreaded self-attention layer and MLP layer.

The input text \( x_q \in \mathbb{R}^{L \times |V|} \) is embedded to \( \tilde{x}_q \in \mathbb{R}^{L \times H} \) with a word embedding matrix \( T \in \mathbb{R}^{|V| \times H} \) and a position embedding matrix \( T^{\text{pos}} \in \mathbb{R}^{(L+1) \times H} \). Here, $L$ represents the input sequence length, $V$ represents the vocabulary size, and $H$ represents the embedding dimension size. 

The input image \( x_i \in \mathbb{R}^{C \times H_t \times W} \) with $C$ being the channel, $Ht$ and $W$ being the height and width is sliced into patches and flattened to \( v \in \mathbb{R}^{N \times (P^2 \cdot C)} \) where \( (P, P) \) is the patch resolution and \( N = Ht \times W/P^2 \) with $N$ signifying total number of patches. Followed by linear projection \( V \in \mathbb{R}^{(P^2 \cdot C) \times H} \) and position embedding \( V_\text{pos} \in \mathbb{R}^{(N+1) \times H} \), \( v \) is embedded into \( \tilde{v} \in \mathbb{R}^{N \times H} \).

\begin{figure}[hb]
    \centering
    \includegraphics[scale=1]{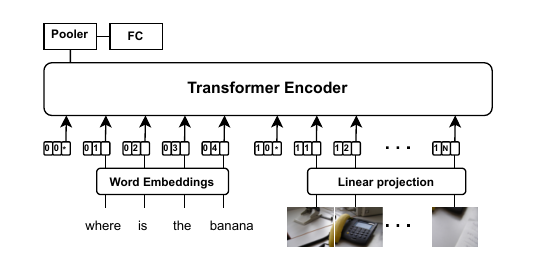}
    \caption{Surrogate Architecure inspired from \cite{dosovitskiy2020image}}
    \label{fig:surrogate}
\end{figure}

The text and image embeddings are summed with their corresponding modal-type embedding vectors \( t^{\text{vtx}}, v^{\text{vtx}} \in \mathbb{R}^H \), then are concatenated into a combined sequence \( z^0 \). The contextualized vector is iteratively updated through \( D \)-depth transformer layers up until the final contextualized sequence. A pooled representation of the whole multimodal input, and is obtained by applying linear projection.  Figure \ref{fig:surrogate} represents the surrogate architecture.  In the positional embedding, the first element represents the modal-type embedding.  For text, the second position represents the token position embedding, and for image, the second position represents the patch position. $*$ represents the extra learnable embedding. 

\subsubsection{Surrogate Alignment}
The goal is to align the surrogate architecture that takes in an image and text and generates a joint representation.  
To align the representations $r_{iq}$ and $r_a$, we adopt a contrastive loss to encourage the representations to be closer. Specifically, we aim to minimize the cosine similarity between the positive pair $(r_{iq}, r_a)$ while maximizing the cosine similarity between the negative pairs $(r_{iq}, r_a')$ and $(r_{iq}', r_a)$, where $r_a'$ and $r_{iq}'$ are negative samples from the batch. The contrastive loss is defined as:

\begin{equation}
\mathcal{L} = -\log \left( 
\frac{ps(r_{iq}, r_a)}{
ps(r_{iq}, r_a) + \sum_i ps(r_{iq}, r'_{ai}) + \sum_j ps(r'_{iqj}, r_a)
} \right)
\end{equation}


where $ps$ represents $\exp(\cos(x,y))$

Where $\cos(x, y)$ is the cosine similarity between vectors $x$ and $y$. By minimizing this loss, we encourage the multimodal and text representations of the correct $(x_i, x_q, x_a)$ (image, question, answer) triplets to be pulled closer together in the embedding space.  We fine-tuned the surrogate architecture on two training sets, MSCOCO and Flicker30k dataset.

\subsection{Coordinated Attack}

In this section, we will discuss the attack strategy for the individual modules.  
Algorithm \ref{algorithmone} represents the overall flow of generating the adversarial samples.  Once the adversarial samples are generated, they are evaluated on victim models to see the change in the output. We use the same gradients from a defined objective to generate adversarial perturbations for both the image module and the text module making attack more unified.

\subsubsection{Image perturbations}

Our system employs a Surrogate encoder that has been aligned to generate adversarial perturbations, to reduce the similarity between the embeddings from the surrogate model and the answer model. This section elucidates the methodology to craft adversarial samples for the image input, thereby fooling a victim model \( V \).


Given a question $x_q$, an image $x_i$, and an answer $x_a$, the encoders generate embeddings \( E_{iq}(x_q, x_i) \) and \( E_{a}(x_a) \), respectively. The similarity between these embeddings is measured using cosine similarity. The objective of the adversarial attack is to reduce this similarity between the embeddings by adding adversarial perturbations thereby deceiving the model.  Equation \ref{cosinesim} represents the cosine similarity between the two embeddings. 


\begin{equation}
    \mathcal{L}(x_q, x_i, x_a) = \frac{E_{iq}(x_q, x_i) \cdot E_{a}(x_a)}{\|E_{iq}(x_q, x_i)\| \|E_{a}(x_a)\|}
    \label{cosinesim}
\end{equation}

The attack strategy involves iteratively perturbing the image to minimize the cosine similarity between \( E_{iq}(x_q, x_i) \) and \( E_{a}(x_a) \) to generate $x_i'$.  


where \( x_i' = x_i + \delta \) represents the perturbed image and \( \delta \) is the adversarial perturbation applied to the original image \( x_i \).
We apply an iterative method used to optimize the adversarial perturbation \(\delta\) such that it minimizes the objective function \(\mathcal{L}(x_i', q, a)\). The process starts with an initial perturbation \(\delta_0 = random\_bounded \_initialization\).

For each iteration \( t = 0, 1, \ldots, T-1 \), the perturbed image along with the question is used to compute the loss using the cosine similarity. The gradient of the loss function \(\mathcal{L}\) with respect to the perturbed input of the previous step is calculated using backpropagation. The sign of this gradient is used to iteratively update the perturbation, ensuring that each step moves in the direction that maximally reduces the cosine similarity between the embeddings.  The perturbation after each step is represented as,

  \begin{equation}
  \footnotesize
       \delta_{t+1} = \Pi_{\epsilon} \left( \delta_t + \alpha \cdot \text{sign} \left( \nabla_{\delta} \left( -\frac{E_{iq}(q, x_i + \delta_t) \cdot E_{A}(a)}{\|E_{iq}(q, x_i + \delta_t)\| \|E_{A}(a)\|} \right) \right) \right)
  \end{equation}

The perturbed image that has caused the lowest cosine similarity after \( T \) iterations is \( x_i' = x_i + \delta_T \).  $\alpha$ represents the step size,

\(\nabla_{\delta} \mathcal{L}\) represents the gradient of the loss function with respect to the perturbation \(\delta\).
   - \(\Pi_{\epsilon}\) is the projection operator ensuring the perturbation remains within the \(\epsilon\)-ball around the original image \( x_i \):
     \[
     \Pi_{\epsilon}(\delta) = \text{clip}(\delta, -\epsilon, \epsilon)
     \]
The projection operator \(\Pi_{\epsilon}\) ensures that the perturbation \(\delta\) satisfies the constraint \(\|\delta\|_{\infty} \leq \epsilon\), keeping the perturbation imperceptible.  For our experiments, we maintained the $\epsilon = 8/255$.

\subsubsection{Text Attack}

To generate adversarial examples that minimize the cosine similarity between question and answer embeddings and instead of searching for a single adversarial example, we aim to find a distribution of adversarial questions $P_\Theta(x_q)$ parameterized by $\Theta$, such that when sampling $\tilde{x}q \sim P_\Theta(x_q)$, the cosine similarity between the embeddings $E_{iq}(\tilde{x}_q, x_i)$ and $E(x_a)$ of the adversarial question and original answer is minimized.



To instantiate the adversarial distribution $P_\Theta(x_q)$, we leverage the Gumbel-Softmax technique \cite{jang2016categorical} which provides a simple way to sample from a categorical distribution while maintaining differentiability. Let $\tilde{\pi}_1, ..., \tilde{\pi}n$ be samples from $\tilde{P}\Theta$, the Gumbel-Softmax distribution with temperature $\tau$ parameterized by $\Theta \in \mathbb{R}^{n \times V}$, which draws samples $\pi$ by independently sampling where $V$ is the vocabulary size and $n$ is the sequence length. Each $\tilde{\pi}_i \in \mathbb{R}^V$ is a vector representing a categorical distribution over vocabulary tokens at position $i$. We define the adversarial question $\tilde{x}_q = e(\tilde{\pi}_1) \oplus ... \oplus e(\tilde{\pi}_n)$, the sequence formed by looking up and concatenating the embeddings $e(\cdot)$ of the sampled token distributions.
The objective is to minimize the cosine similarity between the joint question and image embedding $E_{iq}(\tilde{x}_q, x_i)$ and the given answer embedding $E(x_a)$.  The cosine similarity which is the objective is same as equation 2 with both the image and the question perturbed as $\Tilde{x_i}$ and $\Tilde{x_q}$. 


Additionally, to ensure the generated adversarial questions remain fluent and semantically preserving, we incorporate two additional constraints. The first promotes fluency by minimizing the negative log-likelihood $\text{NLL}_g(\tilde{x}_q)$ of the adversarial text under an external language model $g$. The second controls semantic divergence by minimizing $\rho_g(x_q, \tilde{x}_q)$ based on the BERTScore \cite{zhang2019bertscore} which measures the semantic similarity between the original question $x_q$ and adversarial $\tilde{x}_q$ using contextualized embeddings from $g$.
The full objective is a weighted combination:
$$\mathcal{J}(\Theta) = \mathcal{L}(\Theta) + \lambda_{lm} \text{NLL}_g(\tilde{x_q}) + \lambda{sim} \rho_g(x_q, \tilde{x}_q)$$
Where $\lambda_{lm}, \lambda_{sim} > 0$ control the strengths of the language model and semantic similarity constraints respectively.
We optimize $\Theta$ using gradient descent on $\mathcal{J}(\Theta)$ to find the parameters of the adversarial question distribution $P_\Theta(x_q)$. At inference time, we can efficiently sample $\tilde{x}q \sim P\Theta(x_q)$ and input it to the QA system. The sampled $\tilde{x}_q$ will have minimized cosine similarity to the given answer embedding $g(x_a)$, while being fluent and preserving semantic similarity to the original question $x_q$ as guided by the constraints.
This distributional adversarial attack framework provides a powerful and general approach compared to previous heuristic word replacement methods. By leveraging gradient-based optimization on a continuous distribution over the input space, along with differentiable constraints, it can navigate the landscape more effectively to find stronger and more natural adversarial examples that are not hand crafted or hard set.

\begin{algorithm}
\caption{High-Level flow of adversarial perturbation generation for image and text modality}
\label{algorithmone}
\begin{algorithmic}[1]
\STATE \textbf{Input:} Clean image $x_i$, question $x_q$, answer $x_a$, victim model $V$, encoders $E_{iq}$ and $E_A$, perturbation bounds $\epsilon_i$ and , step size $\alpha$, number of iterations $T$, language model $g$, BERTScore function $\rho_g$, weight parameters $\lambda_{lm}$ and $\lambda_{sim}$

\STATE Initialize perturbations $\delta_i = \mathbf{0}$, $\Theta \in \mathbb{R}^{n \times V}$ with small random values
\FOR{$t = 0$ to $T-1$}
    \STATE Sample adversarial question $\tilde{x}_q \sim P_\Theta(x_q)$ using Gumbel-Softmax
    \STATE Compute joint embedding $\mathbf{r}_{iq} = E_{iq}(x_i + \delta_i, \tilde{x}_q)$
    \STATE Compute answer embedding $\mathbf{r}_a = E_A(x_a)$
    \STATE Compute cosine similarity loss:
    \[
    \mathcal{L}_{iq} = \frac{\mathbf{r}_{iq} \cdot \mathbf{r}_a}{\|\mathbf{r}_{iq}\| \|\mathbf{r}_a\|}
    \]
    
    \IF{$\mathcal{L}_i < \text{best\_similarity}$}

        \STATE Update best perturbation for image: $\delta_i^* = \delta_i$

        \STATE Update best perturbation for text: $\Theta^* = \Theta$

        \STATE Update best similarity: $\text{best\_similarity} = \mathcal{L}_i$

    \ENDIF
    
    \STATE Compute full objective for text module:
    \[
    \mathcal{J}(\Theta) = \mathcal{L}_{iq} + \lambda_{lm} \cdot \text{NLL}_g(\tilde{x}_q) + \lambda_{sim} \cdot \rho_g(x_q, \tilde{x}_q)
    \]
    \STATE Update $\Theta$ using gradient descent:
    \[
    \Theta \leftarrow \Theta - \alpha \cdot \nabla_{\Theta} \mathcal{J}(\Theta)
    \]

    \STATE Compute gradient of the loss w.r.t. $\delta_i$:
    \[
    \nabla_{\delta_i} \mathcal{L}_{iq} = \nabla_{\delta_i} \left( -\frac{\mathbf{r}_{iq} \cdot \mathbf{r}_a}{\|\mathbf{r}_{iq}\| \|\mathbf{r}_a\|} \right)
    \]
    \STATE Update perturbation $\delta_i$:
    \[
    \delta_i = \Pi_{\epsilon_i} \left( \delta_i + \alpha \cdot \text{sign}(\nabla_{\delta_i} \mathcal{L}_{iq}) \right)
    \]
\ENDFOR
\STATE \textbf{Output:} Adversarial image $x_i' = x_i + \delta_i$ and adversarial question $\tilde{x}_q \sim P_\Theta(x_q)$
\end{algorithmic}
\end{algorithm}

\begin{table*}[]
\caption{Comparison of the proposed method with State-of-the-art competitors for visual question and answering task on VQA dataset. All results are displayed by Average Success Rate ($\%$).}
\label{tab:one}
\begin{tabular}{l|lll|ll|lll}
\hline
\hline
\multicolumn{1}{c|}{\multirow{2}{*}{Pre-trained Model}} & \multicolumn{3}{c|}{Image Only}                             & \multicolumn{2}{c|}{Text only} & \multicolumn{3}{c}{Multi-modality}                                             \\ \cline{2-9} 
\multicolumn{1}{c|}{}                                   & \multicolumn{1}{l|}{SSP} & \multicolumn{1}{l|}{FDA} & BSA   & \multicolumn{1}{l|}{BA} & RR   & \multicolumn{1}{l|}{CO-Attack} & \multicolumn{1}{l|}{VLAttack} & Ours           \\ \hline
ViLT                                                     & 50.36                    & 29.27                    & 65.20 & 17.24                   & 8.69 & 35.13                          & 78.05                         & \textbf{94.3}  \\ \hline
BLIP                                                     & 11.84                    & 7.12                     & 25.04 & 21.04                   & 2.94 & 14.24                          & 48.78                         & \textbf{91.0}  \\ \hline
GIT                                                      & -                        & -                        & -     & -                       & -    & 51.16                          & 78.82                         & \textbf{80.43} \\ \hline \hline
\end{tabular}
\end{table*}

\begin{table*}[]
\caption{Comparison of the proposed method with competitors for visual reasoning dataset.  All results are displayed by Average success rate}
\label{tab:two}
\begin{tabular}{l|lll|ll|lll}
\hline
\hline
\multicolumn{1}{c|}{\multirow{2}{*}{Pre-trained Model}} & \multicolumn{3}{c|}{Image Only}                             & \multicolumn{2}{c|}{Text only}  & \multicolumn{3}{c}{Multi-modality}                                             \\ \cline{2-9} 
\multicolumn{1}{c|}{}                                   & \multicolumn{1}{l|}{SSP} & \multicolumn{1}{l|}{FDA} & BSA   & \multicolumn{1}{l|}{BA} & RR    & \multicolumn{1}{l|}{CO-Attack} & \multicolumn{1}{l|}{VLAttack} & Ours           \\ \hline
ViLT                                                     & 21.58                    & 35.13                    & 52.17 & 32.18                   & 24.82 & 40.04                          & 66.65                         & \textbf{73.32} \\ \hline
BLIP                                                     & 6.88                     & 10.22                    & 27.16 & 33.8                    & 16.92 & 8.70                           & 52.66                         & \textbf{58.45} \\ \hline
GIT                                                      & -                        & -                        & -     & -                       & -     & 18.66                          & 41.78                         & \textbf{54.54} \\ \hline
\hline
\end{tabular}
\end{table*}

\section{Experiments}
\subsection{Exerimental Setup}

Experiments are conducted on two different datasets VQA dataset and visual reasoning dataset.  We sample 1000 samples randomly from the validation set of the above mentioned dataset.  Each selected sample is correctly classified by all the target models to be considered for the evaluation. The generated adversarial samples were evaluated on 3 different models, ViLT, GIT, BLIP \cite{kim2021vilt, liu2023improved, dai2024instructblip}.
The experiments were conducted on 3 pre-trained VL models and the use of Attack Success Rate (ASR) to evaluate the performance. We evaluate on two different datasets, VQA dataset \cite{goyal2017making} and the visual reasoning dataset \cite{suhr2018corpus}.  For our baselines, we compare our performance with several uni-modal approaches and multi-modal approaches from the recent literature SSP \cite{naseer2020self}, FDA \cite{ganeshan2019fda}, BA \cite{li2020bert}, RR \cite{xu2021r}, Co-Attack \cite{zhang2022towards}, VLA-attack \cite{yin2023vlattack}.  


Experimental details: For all experiments, our surrogate architecture is composed of weights from ViT-B/32 pre-trained on ImageNet, hence the name ViLT-B/32.  Hidden size H is 768, layer depth D is 12, patch size P is 32, MLP size is 3072 and the number of attention heads is 12.  For the answer encoder $E_a$, we use a bert based architecture with 12 transformer blocks, and hidden size as 768, 12 self-attention heads, and a feed-forward network size of 3072.  During the attack we use adam optimizer to compute gradients with respect to our corresponding inputs with a learning rate set to 0.0005.
Our training experiments were conducted on an Ubuntu machine with 8 Tesla V100S-PCIE-32GB GPUs and an Intel Xeon Gold 6246R CPU @ 3.40GHz with 16 cores. Training was distributed across multiple GPUs for surrogate alignment.

\subsection{Details on victim models}

ViLT has proven to have performed well in several downstream tasks. Given an input image $I \in \mathbb{R}^{H \times W \times 3}$ and a sentence T, ViLT yields $M$ image tokens using a linear transformation on the flattened image patches, where each token is a 1D vector and $M = \frac{HW}{P^2}$ for a given patch resolution $(P, P)$. By attending visual and text tokens and a special token $\langle cls\rangle$ in a Transformer encoder with twelve layers, the output feature from the $\langle cls\rangle$ token is fed into a task-specific classification head for the final output. 

Instruct-BLIP is a vision-language instruction tuning framework that enables general-purpose model to solve a wide variety of visual language tasks.  Instruct-BLIP performs vision-language instruction tuning for zero-shot evaluation.  The work uses instruction-aware visual feature extraction that enables flexible feature extraction according to given instructions by providing textual instruction to both the frozen language model and the Q-former, allowing it to extract instruction-aware features from the frozen image encoder. 

GIT \cite{wang2022git} Generative Image-to-text Tranfromer unifies vision-language tasks such as image/video captioning and question answering.  The method simplifies the architecture as one image encoder and one text decoder under a single language modeling task.

\section{Results and Discussion}
In this section, we compare the performance of our proposed method with competitors on the VQA and visual reasoning datasets evaluated on average success rate across several recent pretrained models.  Table \ref{tab:one} shows our result on the visual question and answering task, where our method consistently outperforms other attack strategies in the multi-modality category.  For the ViLT model, our proposed method achieves an ASR of 94.3 percent which is 21 percent more than the most recent state-of-the-art VLAttack and 59 percent more than Co-attack which was released in the year 2022.  With the BLIP model, we reach a success rate of 91 percent, and for GIT with a success rate of 80.4 percent. 

Table \ref{tab:two} presents similar trends for the visual reasoning dataset.  For the ViLT model, we achieve an ASR of 73.32 percent which is 10.01 percent higher than the VLAttack and 33 percent higher than the Co-attack.  With the BLIP we are 10.99 percent more than VLAttack and 49.75 percent more than co-attack.  The results indicate the superiority of our proposed method in handling multi-modality attacks across both datasets. The performance trends highlight the effectiveness of leveraging both visual and textual information for robust VQA and visual reasoning downstream tasks.

\section{Broader Impact and limitations}
The development of multi-modal foundational models has transformed many sectors such as healthcare, finance and many others.  The unique ability of multi-modal models to accept more than one modality give more opportunities for effective and easy interaction.  This work probes these models with small crafted perturbations which completely misleads models that have billions of parameters and are trained on humongous data.  By building such attack strategies helps to expose the model vulnerabilities as well as generate more samples that can be efficiently utilized to fine-tune these models to improve their robustness.  One of the limitation with surrogate based attacks is that, they often rely on synthetic perturbations that may not reflect the real-world perturbations, demanding more evaluation in that direction.

\section{Conclusion}
In conclusion, this research presents a novel coordinated attack strategy tailored for vision-language models, addressing the unique vulnerabilities posed by their multimodal nature. By aligning a surrogate model's responses with those of a text encoder, we establish a foundation for generating adversarial examples across both visual and textual modalities. Our approach, distinct from existing methods, leverages a generic surrogate model capable of crafting adversarial samples without specific victim model feedback, thereby demonstrating its versatility and applicability across various state-of-the-art vision-language models.  Through experimentation on benchmark datasets and evaluation against multiple victim models, our method showcases superior performance in compromising the robustness of vision-language models compared to existing multi-modal and single-modality attack techniques. Our results underscore the effectiveness of our coordinated approach in generating adversarial perturbations that induce disparate model outputs while minimizing similarity between joint representations and corresponding answers.

{
    \small
    \bibliographystyle{ieeenat_fullname}
    \bibliography{main}
}


\end{document}